\documentclass[cmbright]{WileySTAT-V1} 

\usepackage{natbib}     

\usepackage{dcolumn}    
\newcolumntype{d}[1]{D{.}{.}{#1}}

\def\volumeyear{2018}
\def\volumenumber{99}

\runninghead{First Author and Second Author}{Short title}

\theoremstyle{plain}
\usepackage{tablefootnote} 

\usepackage[acronym]{glossaries}

\newacronym{ml}{ML}{machine learning}
\newacronym{llm}{LLM}{large language model}
\newacronym{sota}{SotA}{state-of-the-art}
\newacronym{nlp}{NLP}{natural language processing}
\newacronym{ai}{AI}{artificial intelligence}
\newacronym{dl}{DL}{deep learning}
\newacronym{rl}{RL}{reinforcement learning}
\newacronym{gai}{GenAI}{Generative AI}
\newacronym{cot}{CoT}{chain-of-thought}
\newacronym{asi}{ASI}{artificial superintelligence}
\newacronym{agi}{AGI}{artificial general intelligence}
\newacronym{laws}{LAWS}{lethal autonomous weapon systems}
\newacronym{av}{AV}{autonomous vehicle}
\newacronym{nasa}{NASA}{National Aeronautics and Space Administration}
\newacronym{nn}{NN}{neural network}
\newacronym{ann}{ANN}{artificial neural network}
\newacronym{bdi}{BDI}{Belief-Desire-Intention}
\newacronym{oecd}{OECD}{Organisation for Economic Co-operation and Development}
\newacronym{rlhf}{RLHF}{reinforcement learning from human feedback}
\newacronym{prisma}{PRISMA}{Preferred Reporting Items for Systematic Reviews and Meta-Analyses}
\newacronym{wos}{WoS}{Web of Science}
\newacronym{qa}{QA}{question answering}
\newacronym{ie}{IE}{information extraction}
\newacronym{imdb}{IMDb}{internet movie database}
\newacronym{ocr}{OCR}{optical character recognition}
\newacronym{ner}{NER}{named entity recognition}
\newacronym{sa}{SA}{sentiment analysis}
\newacronym{mt}{MT}{machine translation}
\newacronym{asr}{ASR}{automatic speech recognition}
\newacronym{stt}{STT}{speech-to-text}
\newacronym{pos}{PoS}{part-of-speech}
\newacronym{tlp}{TLP}{technical language processing}
\newacronym{tc}{TC}{text classification}
\newacronym{mlp}{MLP}{materials language processing}
\newacronym{ir}{IR}{information retrieval}
\newacronym{lmm}{LMM}{large multimodal model}
\newacronym{lda}{LDA}{latent dirichlet allocation}
\newacronym{bert}{BERT}{bidirectional encoder representations from transformers}
\newacronym{lib}{LiB}{Li-ion battery}
\newacronym{kg}{KG}{knowledge graph}
\newacronym{soc}{SoC}{state-of-charge}
\newacronym{soh}{SoH}{state-of-health}
\newacronym{sse}{SSE}{solid-state electrolytes}
\newacronym{bilstm}{BiLSTM}{bidirectional long short-term memory}
\newacronym{bms}{BMS}{battery management system}
\newacronym{rmse}{RMSE}{root mean square error}
\newacronym{hwfet}{HWFET}{highway fuel economy test cycle}
\newacronym{la92}{LA92}{California Unified Cycle}
\newacronym{udds}{UDDS}{urban dynamometer driving schedule}
\newacronym{us06}{US06}{high acceleration aggressive driving schedule}
\newacronym{dst}{DST}{dynamic stress test}
\newacronym{fuds}{FUDS}{federal urban driving schedule}
\newacronym{dpp}{DPP}{digital product passport}
\newacronym{dbp}{DBP}{digital battery passport}
\newacronym{mcp}{MCP}{model context protocol}
\newacronym{moe}{MoE}{mixture of experts}
\newacronym{fair}{FAIR}{Findability, Accessibility, Interoperability, and Reusability}
\newacronym{sei}{SEI}{solid electrolyte interphase}
\newacronym{cei}{CEI}{cathode electrolyte interphase}
\newacronym{t5}{T5}{text-to-text transfer transformer}
\newacronym{sib}{SiB}{Sodium-ion battery}
\newacronym{rnn}{RNN}{recurrent neural network}
\newacronym{ess}{ESS}{energy storage system}
\newacronym{res}{RES}{renewable energy system}
\newacronym{mdkg}{MA-DKGCN}{Multi-Aspect Dynamic Knowledge Graph Convolutional Network}
\newacronym{pmc}{PMC}{policy model consistency}
\newacronym{certj}{CE-RTJE}{candidate entity-based relational triple joint extraction}
\newacronym{ieeex}{IEEE Xplore}{Institute of Electrical and Electronics Engineers Xplore}
\newacronym{shap}{SHAP}{Shapley additive explanations}
\newacronym{gpt}{GPT}{generative pretrained transformer}
\newacronym{nlu}{NLU}{natural language understanding}
\newacronym{nlg}{NLG}{natural language generation}
\newacronym{ev}{EV}{electric vehicle}
\newacronym{tm}{TM}{topic modeling}
\newacronym{eu}{EU}{European Union}
\newacronym{rte}{RTE}{relational triple extraction}
\newacronym{ad}{AD}{abbreviation detection}
\newacronym{od}{OD}{ontology development}
\newacronym{rer}{RER}{recipe extraction and retrieval}
\newacronym{dg}{DG}{dialog generation}

\begin{document}

\title{From the Rock Floor to the Cloud: A Systematic Survey of State-of-the-Art \acrshort{nlp} in Battery Life Cycle}

\author{Tosin Adewumi\affil{k}\corrauth
Martin Karlsson,\affil{b} Marcus Liwicki,\affil{k} Mikael Sjödahl,\affil{c} Lama Alkhaled,\affil{k} Rihab Gargouri,\affil{k} and
Nudrat Habib,\affil{k}}

\address{%
\affilnum{k}Machine Learning Group, Luleå University of Technology (LTU), Sweden \\
\affilnum{b}Research Center for Advanced Battery Technology, LTU, Sweden.\\
\affilnum{c}Experimental Mechanics, Luleå University of Technology (LTU), Sweden.
}

\corremail{tosin.adewumi@ltu.se}

\begin{abstract}
This work presents a systematic review of the application of \acrfull{nlp} along the entire battery life cycle, instead of the review of one stage or method, and introduces a novel conceptual \acrfull{tlp} framework for the \acrfull{eu}'s proposed \acrfull{dbp} and other general battery predictions, as the main objective.
It follows the \acrfull{prisma} method and employs three reputable databases or search engines, including Google Scholar, \acrfull{ieeex}, and Scopus, searched in October, 2025.
The search criteria are 'text', 'language processing', and 'battery' while the exclusion criteria are non-relevant papers and those published before 2017, which was applied to all databases, except \acrshort{ieeex} because it turned up fewer results.
Consequently, it assesses 274 scientific papers before the critical review of the final 66 relevant papers.
We publicly provide artifacts of the review for validation and possible reproducibility. 
The findings show that new \acrshort{nlp} tasks are emerging in the battery domain, which facilitate materials discovery and other stages of the life cycle.
Notwithstanding, challenges remain, such as the lack of standard benchmarks.
The proposed \acrshort{tlp} framework, which incorporates agentic \acrshort{ai} and optimized prompts, may be apt for tackling some of the challenges.

\end{abstract}


\maketitle

\section{Introduction}
\label{sec1}

In the fast-advancing field of \acrfull{nlp} and its many growing application areas, including battery production, new tasks are evolving.
A battery is an \acrfull{ess} that is capable of supplying electrical energy and is usually made of electrochemical substances \citep{mitali2022energy,winter2004batteries}.
Meanwhile, \acrshort{nlp} is a field with methods and tools for machines to analyze, understand and, possibly, generate natural human language \citep{eisenstein2019introduction,liddy2001natural}.
The battery production life cycle, as in many other fields, involves humans who communicate in natural language and store records of some of the communication about the entire complex value chain.
Consequently, as a motivation, this study has the \textbf{main objective of a review of the application of \acrshort{nlp} along the entire value chain of a battery life cycle, from material sourcing to repurposing or end of life}.
The research questions we address are: (1) What are the \acrshort{nlp} tasks along the entire value chain of a battery life cycle and (2) What are some of the datasets, models, and challenges in the application of \acrshort{nlp} to battery research?
Both successful and unsuccessful attempts along the value chain are usually documented in structured or unstructured reports.
The increasing storage demand because of environmental sustainability is leading to increased research and documentation involving the popular \acrfull{lib}, among others, with liquid electrolytes, known for their conductivity, to possible replacement by non-flammable \acrfull{sse} \citep{mahbub2020text,shon2023extracting}.
The documentations create a wealth of information and may include technical manuals, production logs, patents, scientific literature, quality reports, maintenance records, \acrfull{dbp}, and more.
Therefore, \acrshort{nlp} and \acrfull{tlp}, which addresses a specific domain's technical terms \citep{brundage2021technical,dima2021adapting,lowenmark2023labelling}, have the potential to lead to speed-up in battery innovation, including materials discovery, leading to shorter innovation time.

Many factors are important for the improvement of battery production and the understanding of battery behavior.
These include both qualitative and quantitative information on material composition, environmental conditions, understanding of the chemical reactions, and mass transport \citep{lyonnard2025building}.
Experimental methods, which are usually effective, have their drawbacks.
For example, they can be resource-intensive, time-consuming, and involve trial-and-error studies \citep{zuo2025large,mahbub2020text}.
Therefore, researchers have observed that incorporating \acrshort{nlp} in parts of the battery production pipeline can facilitate relevant parts of the value chain, especially as there are thousands of articles on battery research from which to mine \citep{mahbub2020text}.
To address our research questions, we followed the \acrfull{prisma} method \citep{page2021prisma} for systematic survey of the literature between 2017 and 2025 using 3 reputable databases or search engines: Google Scholar, \acrfull{ieeex}, and Scopus.
As a result, we assessed 274 papers before our final list of 66 papers for our findings, where we (1) map the field, (2) compare methods, (3) identify challenges, and (4) introduce a conceptual \acrshort{tlp} framework for battery-related predictions.

\subsection*{Contributions}

This work provides a suitable entry point for newcomers to the field of \acrshort{nlp} applications in the battery domain and an overview of the intersection of \acrshort{nlp} and battery science.
Our main contributions are three-fold and address gaps in the existing literature.
They include:

\begin{enumerate}
    \item We provide, to the best of our knowledge, the first systematic survey of \acrshort{sota} \acrshort{nlp} across the whole value chain of a battery life cycle.
    \item We curate and synthesize important comparative information about the results, models, data, and existing challenges from many different studies, based on the search results.
    \item We introduce a simple and novel conceptual \acrshort{tlp} framework for \acrfull{dbp} and general battery predictions, which combines the capabilities of \acrfull{ai} agents, \acrlong{llm}s (\acrshort{llm}s), and optimized hard and soft prompts.
\end{enumerate}

The rest of this paper is structured as follows.
In the next section (Section \ref{background}), we discuss related work that have surveyed \acrshort{nlp} applications in the battery domain, the basics of \acrshort{nlp}, its traditional tasks, and the battery life cycle, from material sourcing to recycling or repurposing.
In Section \ref{method}, we describe the materials and methods, including the search databases, criteria, and the \acrshort{prisma} method and chart.
In Section \ref{findings}, we present our findings in a clear and structured way, comparing and describing some important works across the value chain.
In Section \ref{discussion}, we discuss ongoing challenges and introduce our conceptual \acrshort{tlp} framework for battery predictions.
Finally, in Section \ref{conclusion}, we conclude with the main insights from the study and possible future work.

\section{Background}
\label{background}

This section is intended to provide background information on related work that have surveyed \acrshort{nlp} applications in the battery domain and the gaps they have.
It provides the basics of \acrshort{nlp} for battery engineers, who may be new to the field, and 11 of its common traditional tasks in relation to the battery domain.
The section also discusses the battery life cycle stages from material sourcing to recycling.

\subsection{Related Work}
\label{related}

There have been several attempts at surveying \acrshort{nlp} in the battery domain.
They are, however, limited to specific stages of the battery life cycle and tasks, as comparatively shown in Table \ref{comp_table}.
The work by \cite{jiang2025applications} reviewed \acrshort{nlp} application in materials science, with a focus on \acrfull{ie} and materials discovery.
They observed that within materials science, \acrshort{nlp} is still in the early stages but the trend towards the development of material-specific pretrained language models is growing.
\cite{zuo2025large} surveyed the applications of \acrshort{llm}s, addressing two questions of what \acrshort{llm}s offer to support battery-related tasks and how more effective \acrshort{llm}s may be developed?
In contrast, our work offers a more inclusive survey by including discriminative models.
In the review by \cite{pievaste2025artificial}, they highlighted macro-scale property prediction for the properties and performance of bulk materials in real-world applications, including multi-physics performance of a battery, as an example.
The survey centered on \acrfull{ml} methods for materials science, including traditional algorithms and deep learning architectures.
\cite{pei2025language} reviewed the literature, examining how \acrshort{nlp} facilitates the understanding and design of novel materials in materials science.
The focus of their review was on the limitations of \acrshort{llm}s, the creation of a materials discovery pipeline, and the potential of \acrfull{gpt} models to synthesize existing knowledge for sustainable materials.
\cite{osaro2025artificial} surveyed materials discovery and chemical synthesis, observing that \acrshort{ai}-driven frameworks use existing compounds to predict new battery electrode materials with optimized properties for conductivity and stability.
\cite{kim2025beginner} reviewed R-based energy forecasting for different energy sources instead of the more common Python programming language, including a brief mention of batteries.
\cite{lin2025systematic} conducted a systematic review of \acrshort{ai} in patent analysis, discussing works that used principal component analysis (PCA) and random forest (RF).

Earlier works, such as \cite{singh2022systematic}, surveyed deep learning approaches in \acrshort{nlp} for battery materials.
They identified classification, \acrshort{ie}, summarization and materials discovery as some of the tasks or applications of \acrshort{nlp} in the battery domain.
In their perspective paper, \cite{zhao2024potential} discussed the potential for insights on battery research from \acrshort{llm}s, with particular attention on fast charging and ChatGPT.
\cite{clark2022toward} focused on the state of ontology development, emphasizing the need for such in battery research and production.
\cite{lee2023natural} conducted a short review on \acrshort{nlp} for materials discovery.
They observed that the number of scientific papers from different databases that are used vary significantly across different research studies, lacking standardization, and sometimes involved the use of \acrfull{ocr} for older documents.
A review of \acrshort{lib} \acrfull{soc} estimation using deep learning was carried out by \cite{liu2022review}, where they classified the methods into two: structured adjustment and unstructured improvement.
They observed that \acrshort{lib} is currently the dominant battery type and identified the usual data acquisition process for training and verification datasets during a hypothetical driving cycle.
Their review was limited to the \acrshort{soc} and some deep neural networks, especially \acrlong{rnn}s (\acrshort{rnn}s).
A more detailed survey along the entire value chain, such as our work, can draw lessons from one stage to benefit other stages of the value chain.

\begin{table}[H] 
\footnotesize 
\caption{Comparison of previous reviews to our work.\label{comp_table}}
\begin{tabular}{p{0.07\linewidth} p{0.22\linewidth}  p{0.4\linewidth} p{0.25\linewidth} }

\toprule
\textbf{Review}	& \textbf{Lifecycle Stage} & \textbf{Task Discussed\textsuperscript{*}}	& \textbf{Comment}\\
\midrule
\cite{zuo2025large}	& sourcing	&	materials discovery & focused only on \acrshort{llm}s \\
\cite{jiang2025applications} & 	sourcing & \acrfull{ie}, materials discovery	&  \\
\cite{pievaste2025artificial}	& sourcing & materials discovery, \acrshort{ie}	& focused only on material lifecycle \\
\cite{pei2025language}	& sourcing	& materials discovery	& focused on \acrshort{llm}s \& material science \\
\cite{osaro2025artificial}	& sourcing, electrode preparation	& materials discovery, chemical synthesis	&  \\
\cite{kim2025beginner}	& usage	& energy forecasting	&  \\
	&	&	&  \\
\cite{lin2025systematic} & assembly & patent analysis, \acrfull{tm} & \\
\cite{singh2022systematic} &	sourcing & materials discovery, \acrshort{ie}, summarization, \acrfull{tc}, \acrfull{ner}, \acrfull{pos} tagging, \acrfull{mt}, \acrfull{dg}, \acrfull{qa}, summarization	&  focused only on materials science\\
\cite{zhao2024potential}	& usage &	\acrshort{ie} & focused only on ChatGPT 
\\
\cite{clark2022toward} & "battery domain" & \acrfull{od}	& focused only on \acrshort{od} \\
\cite{lee2023natural} &	sourcing & materials discovery, \acrshort{ner}	&  \\
\cite{liu2022review} &	usage & \acrfull{soc}	& focused only on \acrshort{soc}  \\
Ours & sourcing, mixing, coating, drying, calendering, slitting, electrode making, winding, die-cutting, stacking, assembly line, drying oven, filling, formation, charge, tester, module assembly, pack assembly, usage, recycling	& materials discovery, \acrshort{tc}, \acrshort{ner}, \acrfull{ie}, \acrfull{sa}, \acrshort{qa}, summarization, \acrfull{ad}, \acrfull{kg}, \acrfull{pmc}, \acrshort{dg}, \acrshort{rte}, \acrshort{od}, \acrfull{rer}, \acrshort{soc}, \acrshort{mt}, \acrfull{asr}, \acrshort{tm} & discussed different models\\
\bottomrule
\end{tabular}
\noindent{\footnotesize{\textsuperscript{*}Level of discussion is uneven. 
}}
\end{table}

\subsection{\acrshort{nlp} Basics}
\label{nlpbasics}

The general \acrshort{nlp} pipeline for solving a task involves the iterative steps of (1) data acquisition, (2) preprocessing (e.g. removal of unwanted characters), (3) tokenization (e.g. splitting into words), (4) model selection, (5) model training, (6) validation and hyper-parameter tuning, (7) testing, and (8) deployment \citep{lane2025natural,srinivasa2018natural}.
The process of model training with \acrshort{nn}s involves an important step of embedding creation, where the tokenized data is transformed into numerical representation (or embeddings) of a lookup table in a relatively small dimension \citep{lane2025natural}.
The \acrshort{nn} model training is based on minimizing a loss function (e.g. cross-entropy loss) to maximize a given utility function by comparing the model predictions to ground truth labels, in supervised learning, and backpropagating to iterate the procedure until a convenient validation loss is obtained so that the model can generalize to unseen data at test time.
The procedure is called unsupervised learning when there are no labels to compare predictions with and it must only learn patterns within the data.
These two paradigms form the ends of the learning spectrum with variants in between, including \acrfull{rl}.
\acrshort{llm}s undergo self-supervised learning as pretraining before undergoing different types of post-training and deployment for inference.

Models have become more complex and deep from the shallow, non-contextual networks of the past, such as Word2Vec \citep{mikolov2013efficient}.
Important concepts from past models can be found in recent models (e.g. embeddings in the early layers of deep models).
Recent \acrfull{sota} models, like \acrshort{llm}s, are based on the \acrshort{sota} Transformer architecture, which originally had both an encoder and decoder \citep{vaswani2017attention}, though recent \acrshort{llm} implementations use only the decoder.
For efficiency, researchers decouple the encoder for discriminative language understanding tasks (e.g. \acrfull{tc}) while the decoder is used for language generation (or generative) tasks (e.g. summarization) though it is also suitable for discriminative tasks.
Examples of discriminative models include \acrfull{bert} and its many variants \citep{devlin2019bert,liu2019roberta,he2021debertav3} while examples of generative models include the \acrshort{gpt} series \citep{brown2020language,openai2024gpt4ocard}.
Some of these models are used in battery-related research and predictions, as will be discussed later.
For example, \acrshort{gpt}4 has been used in chemical \acrfull{ner} and other tasks by
\cite{lee2024text}.
Qwen2.5-7B and Gemma2-9B are \acrshort{llm}s used in \cite{zheng2025cognition}.
Below, we identify some common traditional tasks of \acrshort{nlp}, i.e. those before the advent of \acrshort{llm}s, where the first 6 are categorized as \acrfull{nlu} tasks and the remaining as \acrfull{nlg} tasks, except the last one, which is an unsupervised learning task.
\acrshort{nlu} tasks are usually evaluated with metrics like accuracy or F1 score and \acrshort{nlg} tasks with n-gram-based metrics like ROUGE \citep{lin-2004-rouge,gehrmann-etal-2021-gem} or semantic-based models like \acrshort{bert}Score \citep{bert-score}.
Noteworthy that despite the improvements in \acrshort{nlp}, challenges, such as hallucinations and biases, still exist in the field \citep{pettersson2024generative}.

\subsubsection*{Some Common Traditional \acrshort{nlp} Tasks}

\begin{enumerate}

\item \acrshort{tc}: It is the general activity of automatically labeling natural language texts with thematic categories from a predefined set \citep{sebastiani2002machine}. A variety of standard definitions of \acrshort{tc}  tasks exists in \acrshort{nlp}.
Some examples are \acrfull{sa}, author attribution, news classification, and \acrshort{ner} (for token classification) \citep{gasparetto2022survey,habib2025trends}.
Each of these sub-tasks have many datasets available and they are widely used in research.
In battery research, \acrshort{tc} is used to automatically select battery related documents from a large collection, making literature screening and dataset building faster.
For example, Battery\acrshort{bert} was fine‑tuned on labeled abstracts to classify whether a paper is battery‑related, enabling more focused materials data mining \citep{huang2022batterybert}. 

    \item 
\acrshort{ner}: It seeks to identify and classify mentions of specific entities often referred to as rigid designators into predefined semantic categories such as chemical compounds, persons, locations, organizations, and others \citep{adelani2021masakhaner}.
\acrshort{ner} enables structured extraction of key information from unstructured text, supporting tasks like search, summarization, and knowledge graph construction. 
Commonly used datasets include CoNLL‑2003, OntoNotes 5.0, and W-NUT \citep{hu2024deep,tjong-kim-sang-de-meulder-2003-introduction,derczynski-etal-2017-results}.
\acrshort{ner} can be used to extract chemical and material names from battery literature to support the automatic generation of structured datasets or to identify materials, synthesis descriptions, and phase labels that are relevant to battery research from large collections of scientific abstracts \citep{huang2022batterydataextractor,weston2019named}.

    \item 
Part-of-Speech (\acrshort{pos}) tagging: It is the process of automatically  assigning each word in a sentence a grammatical category such as noun, verb, adjective, or adverb, based on its definition and context within the sentence \citep{toutanova2003feature}. 
It is a core \acrshort{nlp} task that helps models to understand sentence structure, supporting downstream applications like parsing, \acrshort{ner}, and \acrshort{ie}.
Some of the datasets include part of the Penn Treebank and AfricaPOS \citep{dione-etal-2023-masakhapos,marcus-etal-1993-building}.
While \acrshort{pos} tagging improves contextual understanding, it can struggle with word ambiguity, domain-specific terms, and noisy technical text \citep{manning2011part}.

    \item 
\acrshort{ie}: It refers to the automatic identification and organization of specific types of information, like entities, relationships, and events from unstructured or semi-structured text, transforming natural language into structured, machine-readable data \citep{grishman1997information}. \acrshort{ie} can help battery research and production by extracting key data, like capacity, voltage, and efficiency, from text and turning them into structured formats. 
This makes it easier to build battery databases, compare performances, and spot useful trends for material and design choices \citep{huang2022batterydataextractor}.
An example dataset includes POLYIE \citep{cheung-etal-2024-polyie}.

    \item 
\acrshort{sa}: It is a \acrshort{tc} task of automatically detecting and classifying emotions or opinions in text.
This is typically positive and negative or may contain neutral \citep{maas-EtAl:2011:ACL-HLT2011}.
It plays a key role in applications such as customer feedback analysis, social media monitoring, and product review mining.
It enables users to extract information from textual data automatically to support fast decisions about customer satisfaction or product improvement, even within the battery value chain \citep{wankhade2022survey}.
Example datasets include Yelp, \acrfull{imdb}, and movie review datasets \citep{minaee2021deep,maas-EtAl:2011:ACL-HLT2011}.

    \item 
Question Answering (\acrshort{qa}):
It is a task involving systems that can understand a natural language question and automatically return an accurate and relevant answer, either from a given context (Extractive \acrshort{qa}) or based on previously learned knowledge \citep{abdel2023deep}.
This task is particularly useful in education \citep{11364207,pettersson2024generative}.
\acrshort{qa} systems allow users to retrieve precise information quickly without reading full documents, even in the battery domain, making them highly useful for time-sensitive tasks and large-scale knowledge access.
Examples of datasets include  Stanford Question Answering Dataset (SQuAD) 2.0 and Conversational Question Answering Challenge (CoQA) \citep{yatskar-2019-qualitative}.

    \item 
Summarization:
It aims to produce a concise and coherent summary of a document or documents that captures the main ideas of the source text while reducing its length \citep{el2021automatic}.
Text summaries of battery documents can lead to more efficient workflows.
Like with many generative tasks that have different objective metrics to evaluate the task, human evaluation is usually the gold standard, though subjective.
Providing human-generated reference texts for objective metrics is thus essential.
Datasets include BOOKSUM, ArXiv, and PubMed \citep{kryscinski-etal-2022-booksum,cohan-etal-2018-discourse}.

    \item 
\acrshort{mt}: It is a process of automatic conversion of text from one natural language to another \citep{jiang2020natural}. Since this helps in allowing people to access content in different languages without human intervention, even in cross-border partnerships in the battery domain, it is useful in communication and multilingual applications.
While \acrshort{mt} can speed up information access, it's challenging when it concerns complex languages, idioms, and other linguistic issues \citep{wang2022progress,adewumi-etal-2022-potential}.
Datasets include those from the Workshop on Statistical Machine Translation (WMT) and FLORES \citep{guzman-etal-2019-flores,bojar2014findings}.

    \item 
Automatic speech recognition (\acrshort{asr}): This is also known as speech-to-text (STT) and is the process of converting spoken language into written text using computational models \citep{kheddar2024automatic}. 
\acrshort{asr} is useful in applications like voice assistants and transcription services as it helps users interact with machines more naturally. 
Challenges include dealing with background noise, accents, and overlapping speech  \citep{benzeghiba2007automatic}.
Datasets include LibriSpeech and Google-Sc, among others \citep{kheddar2024automatic}.

\item Dialog Generation (\acrshort{dg}): It is the single or multi-turn conversations generated by models communicating with humans or another model \citep{info13060298}.
The task can be divided into task-oriented or open-domain dialog generation.
\acrshort{llm}s are used in open-domain dialog generation, including materials dialog, and are being adapted in task-oriented dialog systems also (e.g. flight bookings) \citep{lee2024investigating}.
Phenomenal improvements in dialog coherence and other features have been witnessed in the past few years in open-domain dialog generation.
However, some limitations like hallucinations still exist \citep{adewumi-etal-2025-limitations}.
A dataset example includes the Ubuntu dialog dataset \citep{info13060298}.

\item Topic modeling (\acrshort{tm}): It learns patterns in the data in an unsupervised manner to discover latent topics and infer topic proportions of documents \citep{wu2024survey}.
It can be used to support other tasks for better results.
Two common challenges with topic modeling are trivial topics, which are based on uninformative words, and repetitive topics, where synonyms exist.
An example of a dataset for the task is 20newsgroup \citep{wu2024survey}.
    
\end{enumerate}

\subsection{Battery Life Cycle}
\label{bat_lifecycle}

\begin{figure}[h!]
\centering
\includegraphics[width=0.95\textwidth]{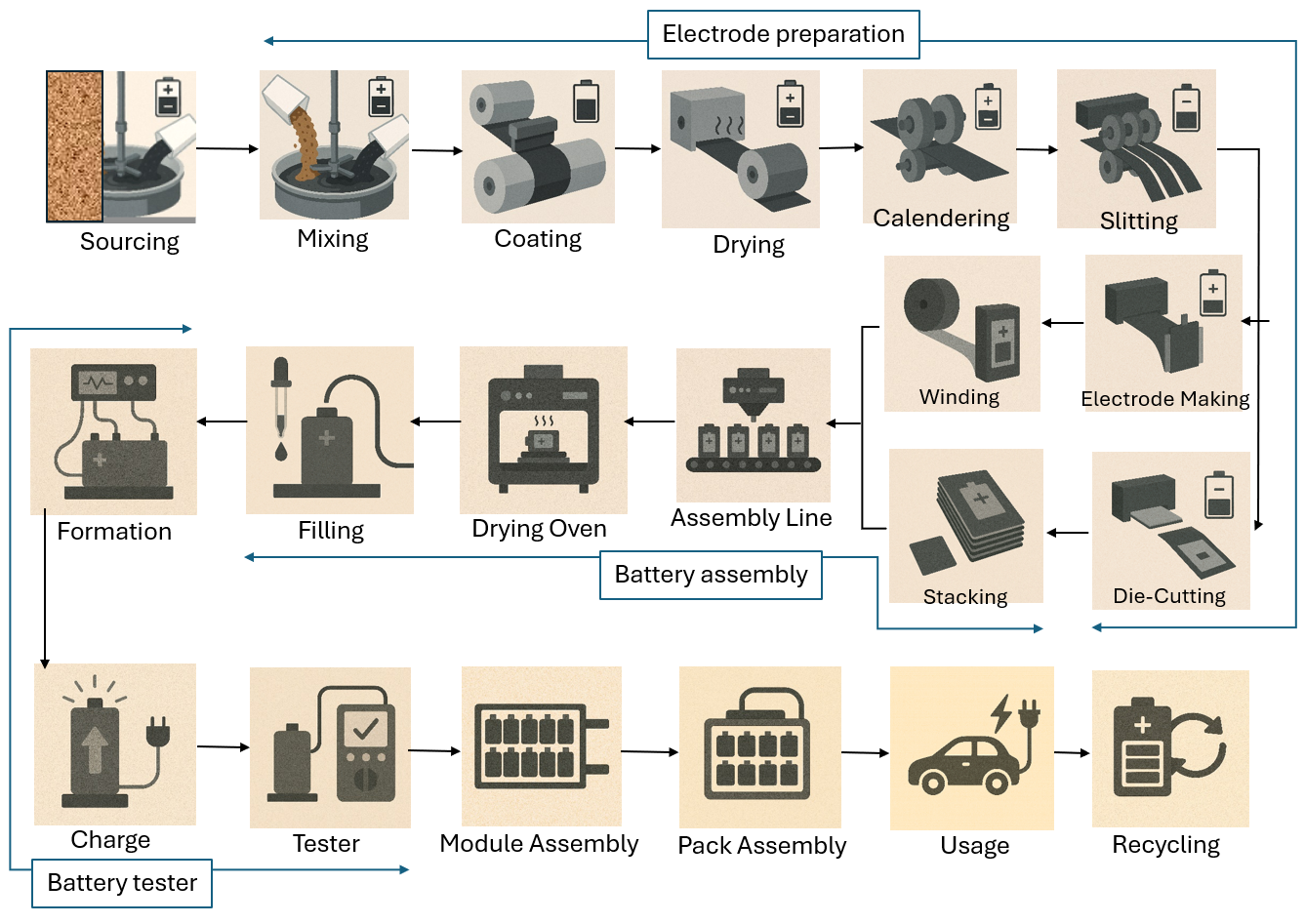}
\caption{Battery life cycle stages synthesized from the literature \citep{xiao2025mining,orum2023lithium,nekahi2025advanced,Lee2025}} (\footnotesize The categorization is not intended as a hard delimitation. Images are \footnotesize{ChatGPT-5-generated}).
\label{batt_prod}
\end{figure}

The increase in \acrshort{res}s and \acrlong{ev}s (\acrshort{ev}s) has made the battery manufacturing industry one of the fastest-growing industries \citep{nekahi2024comparative}. 
The \acrshort{lib} has attracted particular interest because of its high energy, high power density, long cycle life, and potential across electronic mobility and stationary storage.
Battery production is a complex process that involves many steps, from material sourcing, mining and refining raw materials to assembling cells together into finished battery packs \citep{xiao2025mining,orum2023lithium,nekahi2025advanced}.
During every step of the production process, specific physical and chemical processes must be followed, supported by quality assurance and environmental controls. These processes generate large volumes of textual data throughout the entire life cycle, which can be leveraged through \acrshort{nlp} for process analysis, compliance, and optimization \citep{Lee2025,jiang2025applications}.
Figure \ref{batt_prod} depicts the common stages of the battery life cycle, which involve about 17 stages, as synthesized from the literature.
We broadly categorize and discuss each of the stages briefly below.
The categorization is not intended as a hard delimitation.

\subsubsection{Material Sourcing}

\textbf{Material sourcing} is the foundation of the entire production chain. Cobalt, nickel, manganese, graphite, and lithium are key raw materials, each with unique electrochemical properties.
Mining and refining operations are required to achieve the purity levels needed for batteries \citep{xiao2025mining}. 
For example, lithium comes from hard rock mineral deposits (e.g. spodumene, LiAl(SiO$_3$)$_2$) and brines pools (e.g. in South America).
The refining process, such as acid roasting and precipitation, then follows to produce lithium carbonate or hydroxide. 
Material sourcing significantly impacts both environmental outcomes and production costs \citep{degen2023energy}. It also generates extensive documentation, such as mining reports, safety sheets, and procurement contracts.

\subsubsection{Material mixing}
The sources of documentation coupled with the wealth of information on the internet make \acrshort{nlp} tasks like materials discovery, \acrfull{qa}, and \acrfull{ir} through deep analysis of the available information an important component of the previous stage, and possibly the first, to guide informed decisions for \textbf{material mixing}, where conductive additives and polymer binders are homogenously dispersed in a solvent (usually water or N-methyl-2-pyrrolidone) to create a homogenous slurry \citep{Lee2025}.

\subsubsection{Electrode preparation (front stage process)}

Electrode preparation, which converts refined powder into active electrodes, is the most costly and time-consuming stage of the manufacturing process \citep{nekahi2024comparative,orum2023lithium}.
Two main approaches exist to process the electrode fabrication step: solvent-based wet processing (which currently dominates) and solvent-free dry processing.
Dry processing is gaining attention due to its lower energy use and environmentally friendly process.
The main sub-steps are:
\textbf{coating}, in which the slurry is applied onto a metallic foil (aluminum or copper) while maintaining precise control of the thickness, \textbf{drying}, which removes solvent and forms a solid film, representing one of the most energy intensive steps in the process, \textbf{calendering}, where the electrodes are compressed between hard rollers to improve contact and mechanical integrity, \textbf{slitting}, where wide rolls are cut into narrow electrode strips,
\textbf{electrode making and die-cutting}, where electrode making is for producing both cathode and anode sheets and die-cutting is for trimming the coated foils into specific geometries, such as cylindrical, prismatic, or pouch shapes \citep{degen2023energy}.
The process documentations, including mixing ratios, coating logs, temperature records, and inspection reports, serve as rich textual sources for \acrshort{nlp}-driven process optimization and predictive quality control \citep{liu2021feature}.
At the laboratory scale, various coating methods are available, including spray coating, spin coating, dip coating, 
doctor blading, and slot-die coating \citep{degen2023energy}.

\subsubsection{Battery Assembly}
In the assembly stage, electrodes and separators are combined into complete cells \citep{attia2025challenges}.
During \textbf{winding and stacking}, depending on the type, electrodes can be wound (for cylindrical cells) or stacked (for prismatic or pouch cells). The anode, separator, and cathode layers are rolled into a tight 'jelly roll' to ensure uniform ion transport.
Stacking arranges sheets in alternating layers, enabling high packing density and consistent performance \citep{degen2023energy}.
\textbf{Assembly line} ensures electrode tabs are welded to terminal leads, and the assembled cell goes through the \textbf{drying oven} to remove residual moisture, that is, the solvent and water from the cathodes and anodes, respectively.
Finally, \textbf{electrolyte filling} introduces a lithium salt solution into the porous structure in a controlled atmosphere.
Each step produces operational and safety logs that may be analyzed using \acrshort{nlp} to detect anomalies or optimize production sequences.

\subsubsection{Battery tester}

This stage initiates the cell's electrochemical activity through controlled charge and discharge cycles, which generate the \acrfull{sei} and \acrfull{cei}. These layers are critical for cell stability and safety \citep{schomburg2024lithium,weng2023modeling}.
\textbf{Formation} generally occurs over several days under constant current, constant voltage (CCCV) conditions. After formation, cells undergo a series of test, including impedance measurement, self-discharge assessment, and safety evaluations, to verify performance consistency and compliance with quality standards \citep{liu2021feature}.
Generally, at the \textbf{charge} stage, the charging method determines battery capacity \citep{huang2020database}.
Methods include CCCV and the cutoff voltages.
The \textbf{tester} is used to qualify and verify the performance of the battery components to ensure they meet electrical and safety standards.
Logs generated during the formation and testing stages are often stored in digital manufacturing systems, which again can serve \acrshort{nlp} application purposes to improve yield and enable predictive maintenance.

\subsubsection{Assembly II}

Validated cells are combined into modules in a \textbf{module assembly}.
A typical battery module consists of several cells connected in series or parallel.
Multiple modules are then combined in the \textbf{pack assembly}, which, like the module assembly, is equipped with thermal management and electronic control systems \citep{liu2021feature}.
In addition, the pack contains a \acrfull{bms}, cooling components, and safety devices.

\subsubsection{Usage}
Once assembled, the pack undergoes conditioning cycles to verify the functionality of sensors, wiring, and control electronics prior to deployment for \textbf{usage}.
The safe use of \acrshort{lib}s relies on the \acrshort{bms}, which manages several parameters, including \acrshort{soc} of the batteries, for efficient function during the life of the batteries \citep{bian2024exploring}.
The \acrshort{bms} also continuously monitors voltage and temperature to ensure safety and efficiency.
The \acrshort{soc} indicates the capacity left in a \acrshort{lib}, thereby providing information against overcharging or undue discharging.
Methods of estimating \acrshort{soc} may be divided into 4 categories: look-up table, Ampere-hour counting, model-based, and data-driven.
The data-driven method has been gaining attention in recent times \citep{bian2024exploring}.

\subsubsection{Recycling}
\textbf{Recycling} forms the final stage of the battery life cycle, especially for batteries that can not be \textbf{repurposed} into secondary life usage.
Valuable materials such as lithium, nickel, cobalt, and copper are recovered through recycling and reintroduced into the supply chain to support a circular battery economy.
Recycling processes and life cycle assessments produce a wide range of textual data, from dismantling instructions and recovery reports to environmental impact studies and compliance records \citep{Lee2025}.
Batteries that cannot be recycled come to their end of life and must be disposed of properly.

\section{Materials and Methods}
\label{method}

\begin{figure}[h!]
\centering
\includegraphics[width=0.8\textwidth]{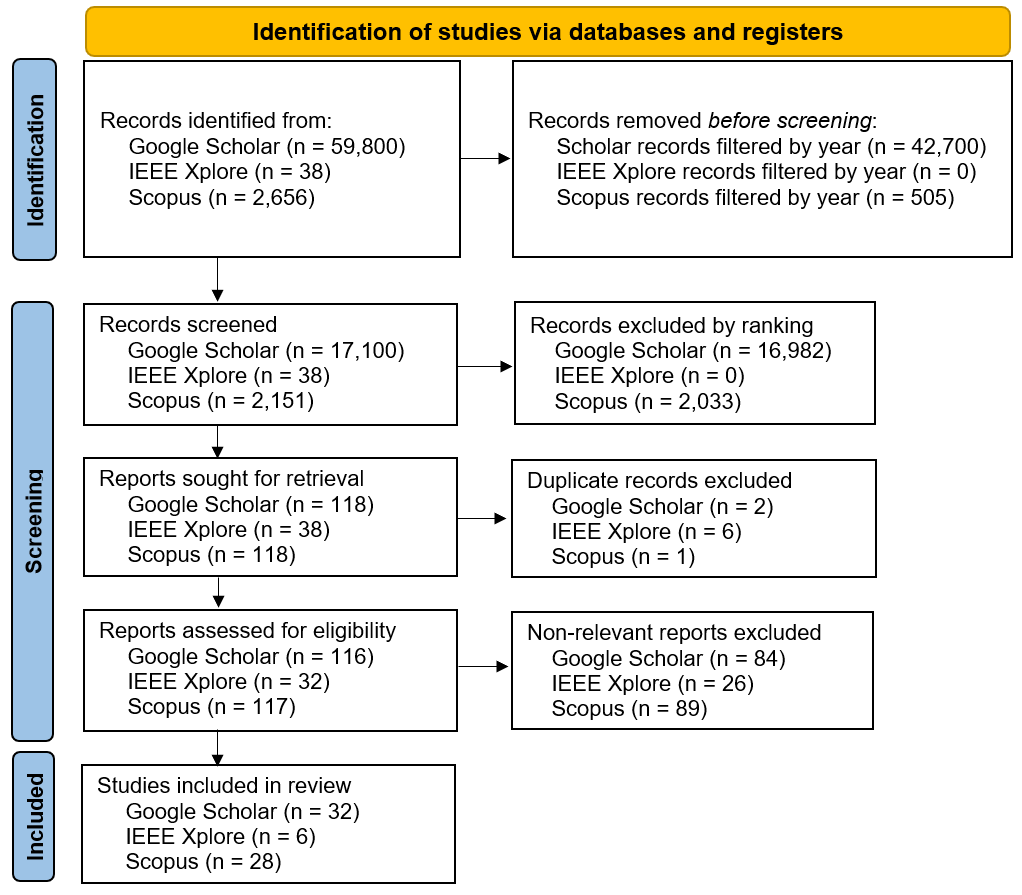}
\caption{\acrshort{prisma} flow diagram for the systematic review.} 
\label{prisma_diag}
\end{figure}

We followed the \acrshort{prisma} method \citep{page2021prisma} for systematic reviews, complying with its guidelines, 
to achieve the stated objective of this work.
Figure \ref{prisma_diag} presents the \acrshort{prisma} flow diagram for the review.
It follows the rigorous and auditable general guidelines recommended for conducting a
systematic literature review \citep{page2021prisma,adewumi2024fairness}.
This review is not registered.
We searched 3 reputable databases or search engines with relevant (inclusion) terms and focused on search results in English.
This  implies the findings are limited in scope to research conducted in English.

\subsection{The Databases}

The databases (or search engines) are Google Scholar, \acrshort{ieeex}, and Scopus.
Noteworthy that they are time-varying and sensitive to ranking algorithms.
They were searched October, 2025.
They are suitable because they index the major publishers or databases.
Each database complements the other.
For example, \acrshort{ieeex} and Scopus serve as additional validation for Google Scholar, which sometimes includes archived (or draft) papers while Scholar has a wider coverage of results compared to the other two.

\subsection{Search Criteria}

The search terms (or keywords) we settled for are 3: 'text', 'language processing', and 'battery'.
These are based on the inclusion criterion of the objective of the study.
This is because \acrshort{nlp} commonly involves written text, 'language processing' applies to \acrlong{nlp}, \acrlong{tlp}, and \acrlong{mlp}, and battery applies to any stage of the battery life cycle.
The search terms were combined using the AND operator on all the 3 databases as "text AND 'language processing' AND battery".
In preliminary search validation, we also attempted 4 combined terms: 'text', 'language', 'processing', and 'battery' but realized this gave a lot more non-relevant results than our earlier terms.
Also, we chose not to include many other terms to avoid losing relevant articles since the objective was to cover the entire life cycle.
Hence, we settled for our initial 3 search terms.
We note that there are numerous possible search terms in \acrshort{nlp} and for this reason we kept it simple by using the 3 terms, which act as catch-all, as much as possible.
For example, “NLP” is an acronym for “natural language processing”, which our search term “language processing” should catch. Similarly, “text mining” should be caught by “text”. Hence, we believe the terms (text, language processing and battery) are very basic for catching many papers (including the sub-tasks) in the search.

\subsection{The \acrshort{prisma} method}

As depicted in Figure \ref{prisma_diag}, the number of search results at every stage of the screening, including the year filter, are provided.
Our choice of year filter (or cut off) from 2025 back to 2017 was because 2017 marked a pivotal period in \acrshort{nlp} with the introduction of the Transformer architecture \citep{vaswani2017attention,huang2022batterydataextractor}.
Using a ranking-based selection procedure, we reviewed the top 118 ranked papers each from Google Scholar and Scopus while we reviewed all the 38 papers from \acrshort{ieeex} since it returned a much smaller number in results.
Hence, as an exception, we did not filter \acrshort{ieeex} by year so as not to lose more results.
Since it’s not practically feasible to review all the thousands of papers in each search, a heuristic that is based on the number of samples from a population was used as a guide for the number 118 and the topmost were selected, based on the reasonable assumption that the most relevant appear at the top.
Screening the titles and abstracts revealed papers that were not relevant.
In situations where it was not immediately clear if a paper was relevant, we performed a quick automatic search for our keywords (especially 'battery') in the body of the paper so the reviewer can read the context around it.
At the full text stage, all relevant papers were reviewed.
Furthermore, we observed that for some non-relevant papers, especially health-inclined ones, the term "battery of ..." or similar appears, where battery is used to mean something different, e.g. a series of things.
The process of identifying duplicate publications involved identifying same titles (or abstracts) across databases.
For validation and reproducibility of our work, the search result links\footnote{search results are limited by being time-varying \& sensitive to ranking algorithms and the truncated papers are a limitation.}, downloadable Scholar list,\footnote{https://github.com/LTU-Machine-Learning/battery\_survey} and the list of non-relevant papers are in the appendix.

\subsection{Use of \acrshort{gai}}
We used \acrshort{gpt}-5 (ChatGPT-5) to generate the images that make up Figure \ref{batt_prod}.
We described each lifecycle stage in prompts and asked it to produce a downloadable png image.
We then combined these, according to what is described in the literature, into a lifecycle diagram as shown in the figure.

\begin{table}[h!]
\small
\centering
\caption{Distribution of research papers across tasks. (\footnotesize Battery prediction\textsuperscript{b} and ontology development\textsuperscript{o} are superscripted while review$_r$ and position$_p$ papers are subscripted. Instances of papers that are identical are indicated with =)}
\label{taxonomy_papers}
\begin{tabular}{p{0.02\linewidth} | p{0.18\linewidth} | p{0.29\linewidth} | p{0.09\linewidth} | p{0.27\linewidth}}
\hline
   \textbf{No.}  & \textbf{Task (\acrshort{nlp} / battery\textsuperscript{b} / ontology\textsuperscript{o} / new\textsuperscript{n})}   & \textbf{Google Scholar} & \textbf{\acrshort{ieeex}} & \textbf{Scopus} \\
      \hline
  1  & \acrshort{tc} & G2$_r$, G7, G8, G23, G25, G31,  & I6 & S5, S29\\
  2  & \acrshort{qa} & G3$_p$, G4, G7, G8, G23 &  & \\

   3  & Materials discovery\textsuperscript{n} & G2$_r$, G9$_r$, G10, G17, G22, G37$_r$, G47$_r$ & I6 & S2, S3$_r$, S7, S43$_r$, S46$_r$, S73$_r$, S78$_r$, S79, S81, S82, S100$_p$ \\

    4 & \acrshort{ie} & G1, G4, G7, G10, G11, G12, G19, G31, G38, G56, G57, G100 &  & S1, S5 \\

    5 &  \acrshort{ner} & G1, G4, G7, G8, G10, G23, G25, G49, G100 & I19 & S1 \\
    6 & Summarization & G25 &  &\\

    7 & Abbreviation detection\textsuperscript{n} & G4, G31 &  & \\

    8 & \acrshort{sa} & G75 & I18, I27 & S13, S18, S19, S53, S55, S67 \\
    9 & Policy model consistency (\acrshort{pmc})\textsuperscript{n} & & & S49 \\

    10 & Relational triple extraction (RTE)\textsuperscript{n} & & & S88 \\
      
    11 & \acrshort{soc} estimation\textsuperscript{b} & G13, G40$_r$ & I1 & \\
    12  & Energy prediction\textsuperscript{b} & & I26 & S91$_r$ \\

    13 & Recipe extraction and retrieval (RER)\textsuperscript{n} & G1, G5, G6 &  &  \\
     
    14 & \acrshort{ir} & G2$_r$, G6, G28 &  & S12, S80 \\
    
     15 & \acrshort{tm} & G1, G18, G89 & I6 & S1, S53, S80, S90, S92, S97\\

    16  & Ontology development\textsuperscript{o} & G15$_r$ & & \\

 \hline
      & Non-relevant papers &  G14, G16, G20, G21, G24, G26, G27, G29, G30, G32, G33, G34, G35, G36, G39, G41, G42, G43, G44, G45, G46, G48, G50, G51, G52, G53, G54, G55, G58, G59, G60, G61, G62, G63, G64, G65, G66, G67, G68, G69=G71, G70=G72, G73, G74, G76, G77, G78, G79, G80, G81, G82, G83, G84, G85, G86, G87, G88, G90, G91, G92, G93, G94, G95, G96, G97, G98, G99, G101, G102, G103, G104, G105, G106, G107, G108, G109, G110, G111, G112, G113, G114, G115, G116, G117, G118  & I2, I3, I4, I5, I7, I8, I9, I10, I11, I12, I13, I14, I15, I16, I17, I20, I21, I22, I23, I24, I25, I28, I29, I30, I31, I32=133=134 =135=136 =137=138  & S4, S6, S8, S9, S10, S11, S14, S15, S16, S17, S20, S21, S22, S23, S24, S25, S26, S27, S28, S30, S31, S32, S33, S34, S35, S36, S37, S38, S39, S40, S41, S42, S44, S45, S47, S50, S51, S52, S54, S56, S57, S58, S59, S60, S61, S62, S63, S64, S65, S66, S68, S69, S70, S71, S72, S74, S75, S76, S77, S83, S84=S63, S85, S86, S87, S89, S93, S94, S95, S96, S98, S99, S101, S102, S103, S104, S105, S106, S107, S108, S109, S110, S111, S112, S113, S114, S115, S116, S117, S118 \\
 \hline
\end{tabular}
\end{table}

\section{Findings: \acrshort{nlp} in the Battery Life Cycle}
\label{findings}

After the careful review of all the relevant papers, we categorized each paper (identified as DatabasePaperNumber, e.g. G2 for Google Scholar paper 2) based on the \acrshort{nlp} tasks they address, resulting in Table \ref{taxonomy_papers} (and the quality of the Scholar papers in Table \ref{quality_papers} in appendix \ref{qual_appendix}).
We synthesized some of the \acrshort{sota} results, where applicable, into Table \ref{sota_results}.
In Table \ref{taxonomy_papers}, there are 16 tasks and some of them are different from the 11 identified in Section \ref{nlpbasics}: 5 are new, 2 are battery-specific, and 1 is ontology development.
Some of the papers address multiple tasks.
In Table \ref{sota_results}, \acrshort{sota} results and models are compared for different datasets per task (with the life cycle stage in the data).
Different metrics are reported, depending on the task, and the category of each model is also identified.
Some of the tasks identified in Table \ref{taxonomy_papers} have no results reported in Table \ref{sota_results} because the relevant paper in the earlier table is either a review paper or does not report a relevant result.
Hence, 11 main tasks are reported in Table \ref{sota_results} and the category of models are deep learning, classical, \acrshort{llm}, and \acrfull{kg}.
In the following subsections, we discuss some details of the relevant papers based on their contributions.

\begin{table}[h!]
\tiny
\centering
\caption{Synthesis of some \acrshort{sota} results. (\footnotesize Comparison across tasks/data is not suitable, coverage across the life cycle stages is uneven} and specific metrics or models are identified with superscripts.)
\label{sota_results}
\begin{tabular}{p{0.02\linewidth} | p{0.12\linewidth} | p{0.29\linewidth} | p{0.19\linewidth} | p{0.2\linewidth}}
\hline
   \textbf{No.}  & \textbf{\acrshort{nlp} task}   & \textbf{Data} (Life cycle stage) & \textbf{Metric score} (F1\% /Accuracy\textsuperscript{a} \% /\acrshort{rmse}\textsuperscript{rm} /ROUGE-1\textsuperscript{RG} /Cosine\textsuperscript{cs}) &  \textbf{Model} (Deep learning /Classical\textsuperscript{c} /\acrshort{llm}\textsuperscript{l} /Knowledge graphs\textsuperscript{kg}) \\      \hline
   1 &  \acrshort{tc} & Battery abstracts (electrode preparation, battery assembly) & 94.47\textsuperscript{a} \citep{huang2022batterybert}  & BatterySci\acrshort{bert} \\
     &  & Battery papers (sourcing) & 96.6\textsuperscript{a}  \citep{choi2024accelerating} & GPT3\textsuperscript{l} finetuned \\
     & & Battery papers (electrode preparation, battery assembly, usage) & 97\textsuperscript{a} \citep{zheng2025cognition} & GPT4\textsuperscript{l} finetuned\\
     & & & & \\
     
  2   & \acrshort{qa} & SQuADv1.1 dev & 89.16 \citep{huang2022batterybert} & Battery\acrshort{bert}-cased \\
     & & Battery device QA (electrode preparation) & 88.21 \citep{choi2024accelerating} & GPT3\textsuperscript{l} finetuned\\

  3 & Materials discovery & SpringerLink \& ScienceDirect (sourcing) & 57\textsuperscript{cs} \citep{he2021prediction} & Word2Vec\textsuperscript{c} \\
     
 4  & \acrshort{ie} & \textit{Unknown} articles (electrode preparation) & 89.58 \citep{gou2024document} &  \textit{Unknown}\\
    & & MKG (sourcing) & 88.27 \citep{ye2024construction} & Darwin\textsuperscript{l} (ER) \\
    & & NLP4SIB (electrode preparation) & 84.1 \citep{munjal2023extracting} & Sci\acrshort{bert}\\
    & & Auto-generated battery materials (sourcing) & 67.98 \citep{huang2020database} & \textit{Unknown}  \\
    & & Battery papers (electrode preparation, battery assembly, usage) & 85 \citep{zheng2025cognition} & Gemma2-9B\textsuperscript{l} \\
    & & & & \\

    
 5  &   \acrshort{ner} & Cell-Assembly (electrode preparaton) & 94.61 \citep{lee2024text} & Battery\acrshort{bert} \\
     & & CHEMDNER (sourcing) & 95.98 \citep{huang2022batterydataextractor} & BatteryOnly\acrshort{bert}-uncased
     \\
     & & Solid-state materials (sourcing) & 95.1\tablefootnote{DSC - sample descriptors category} \citep{choi2024accelerating} & GPT3\textsuperscript{l} finetuned\\
     & & MatScholar (sourcing) & 87 \citep{choudhary2023chemnlp} & XLNet\\
     & & MS-Mention (electrode preparation) & 91.47 \citep{o2021ms} & Sci\acrshort{bert} \\
    & & MS-Mention (electrode preparation) & 91.85 \citep{o2021ms} &  Sci\acrshort{bert}+SOFC \\
    & & MKG (sourcing) & 92.96 \citep{ye2024construction} & Darwin\textsuperscript{l} (ER) \\
    & & manuals (assembly, repurposing) & 93.99 \citep{ren2025automated} &  \acrshort{certj} \\
    & & & &  \\

   6  & Summarization & arXiv-cond-mat (electrode preparation) & 46.5\textsuperscript{RG} \cite{choudhary2023chemnlp} & \acrshort{t5} \\
     
  7 & Abbreviation detection & PLOS & 95.16 \citep{huang2022batterydataextractor} & BatteryOnly\acrshort{bert}-cased\\
     & & \textit{Unknown} articles (electrode preparation) & 92.71 \citep{gou2024document} & \textit{Unknown}\\
     
   8 & \acrshort{sa} & BYD NEVs (usage) & 91.46 \citep{xu2025user} & \acrshort{mdkg}\textsuperscript{kg} \\
     & & AutoHome (usage) & 81 \citep{na5174549decoding} & ALBERT\\
     & & X (formerly Twitter) (usage) & 91.9 \citep{maghsoudi2025novel} &  PyABSA \\
     
   9  & Policy model consistency (\acrshort{pmc}) & Chinese policy documents (assembly, usage) & 7.46\tablefootnote{qualitative metric for consistency, with 10 as the highest score.} \citep{liang5235882quantitative} & ChatGPT\textsuperscript{l} \\
   
   10  & Relational triple extraction (RTE) & manuals (assembly, repurposing) & 95.6 \citep{ren2025automated}  & \acrshort{certj}\textsuperscript{kg} \\
     
 11 & \acrshort{soc} estimation & Panasonic-18650PF (usage) & 1.89\textsuperscript{rm}, 2.21\textsuperscript{rm}, 2.02\textsuperscript{rm}, 2.42\textsuperscript{rm} (\acrshort{hwfet}, \acrshort{la92}, \acrshort{udds}, \acrshort{us06}) at -20\textsuperscript{o}C \citep{bian2024exploring} & ChatGLM-6B\textsuperscript{l} \\
     & & LG-18650HG2 (usage) & 1.89\textsuperscript{rm}, 1.83\textsuperscript{rm}, 1.69\textsuperscript{rm}, 1.48\textsuperscript{rm} at -20\textsuperscript{o}C \citep{bian2024exploring} & ChatGLM-6B\textsuperscript{l} \\
     & & Samsung-18650-20R (usage) & 1.91\textsuperscript{rm}, 2.75\textsuperscript{rm}, 2.72\textsuperscript{rm}, 1.92\textsuperscript{rm} at 0\textsuperscript{o}C \citep{bian2024exploring} &  ChatGLM-6B\textsuperscript{l} \\
     
     & & A123-18650 (usage) & 1.69\textsuperscript{rm}, 2.46\textsuperscript{rm}, 1.86\textsuperscript{rm} (\acrshort{dst}, \acrshort{fuds}, \acrshort{us06} at -10\textsuperscript{o}C) \citep{bian2024exploring} & ChatGLM-6B\textsuperscript{l} \\
 \hline
\end{tabular}
\end{table}

\subsection{Material Discovery and Knowledge Mining}

Apparently, much of the work in the literature and research has gone in the area of material discovery and knowledge mining (such as \acrshort{ie} and \acrshort{ner}), as shown in Table \ref{taxonomy_papers} in terms of the number of papers.
Table \ref{materials_table} comparatively synthesizes some of the findings on materials discovery.
\cite{he2021prediction} conducted unsupervised learning of the literature by using Word2Vec for prediction of solar-chargeable battery materials in materials discovery.
They calculated the cosine similarities (for relevance) between different materials and certain words like "photo-rechargeable".
\cite{choi2024accelerating} studied in-context learning with \acrshort{llm}s in what they call \acrfull{mlp}, which may be considered a subset of \acrshort{tlp}, since battery terms are specific materials terms \citep{huang2022batterydataextractor}.
Battery\acrshort{bert} was introduced by \cite{huang2022batterybert} for the tasks of text classification and extractive \acrshort{qa} by employing continued pretraining from the original \acrshort{bert} weights by \cite{devlin2019bert} before finetuning.
A toolkit that is based on the Battery\acrshort{bert} is BatteryDataExtractor, which was used for the extraction of chemical data \citep{huang2022batterydataextractor}.
Its implementation is, however, limited to two turns of \acrshort{qa} and does not follow a natural conversation style.
\cite{lee2024text} went further and introduced text-to-battery recipe for recipe extraction from the scientific literature.
Recipes contain both materials and instructions about the materials.
\textbf{A major challenge in materials discovery is the lack of a representative metric} or evaluation framework for evaluating successful discoveries without the need for expensive experimentation for validation.
In addition, \acrshort{ie} is a challenging task in the domain because it requires multiple critical variables for materials selection.
To address this challenge, \cite{nie2022automating} proposed a semantic \acrfull{kg} dedicated to \acrfull{lib} cathodes and featuring a dual-attention component that refines word embeddings.
Their implementation is based on the \acrfull{bilstm}.
This is, however, known for limitations like less capacity for long-time dependencies and reliance on serial processing \cite{graves2005bidirectional}, especially compared to the \acrshort{sota} Transformer architecture.

\begin{table}[t]
\footnotesize
\caption{Comparative synthesis of materials discovery \& knowledge mining\label{materials_table}}
\begin{tabular}{p{0.06\linewidth} p{0.43\linewidth} p{0.16\linewidth} p{0.2\linewidth}}
\toprule
\textbf{Study}	& \textbf{Method/Model}	& \textbf{Data}	& \textbf{Limitation}\\
\midrule
\cite{lee2024text} & Text-to-Battery recipe & Cell-Assembly	& Bias towards LiFePO$_4$\\
\cite{zheng2025cognition} &	Cognition-Enhanced Instruction Framework (CEIF) &  Battery papers & Requires several rounds for improvement \\
\cite{huang2022batterybert} &	Battery\acrshort{bert}, BatteryOnly\acrshort{bert}, BatterySci\acrshort{bert} & SQuADv1.1 dev & Size-dependent performance \\
\cite{choi2024accelerating} &	\acrshort{llm}s with in-context learning & Battery papers	& Uncertainty of faithfulness to the original \\
\cite{he2021prediction} &	word2vec with cosine similarity	& SpringerLink \& ScienceDirect			& Based on shallow network\\

\cite{gou2024document} & Document-level information extraction &  Scientific articles	& Requires considerable manual intervention\\
\cite{o2021ms} & Multi-stage annotation & MS-Mentions	& Limited accurate labels\\
\cite{nie2022automating} &	semantic \acrshort{kg}& Materials, Physics \& Chemistry abstracts & Based on shallow skipgram\\
\cite{park2025exploration}  &	Chemeleon using denoising diffusion & Materials Project & Uncertainty of faithfulness of generated compounds\\

\bottomrule
\end{tabular}
\end{table}

Sodium is gaining attention because it is abundant and cheap, leading to \acrfull{sib}, which has similar working principles to \acrshort{lib}.
\cite{gou2024document} performed \acrshort{ie} for cathode materials of \acrshort{sib}.
Their approach combined a number of other related tasks (e.g. chemical \acrshort{ner} and \acrshort{tc}) to improve performance on \acrshort{ie}.
Also, in the chemical \acrshort{ner} contribution by \cite{o2021ms} they annotated labels over a new corpus of
595 synthesis procedures and made the data (MS-Mentions) publicly available.
\cite{bai2025preferable} also used \acrshort{nlp} to guide the screening of single-atom catalysts for the Na-S batteries.
In the work by \cite{zheng2025cognition}, they introduced cognition-enhanced instruction framework (CEIF), where a teacher model provided feedback, prompt refinement, and optimized training data to guide the learning process of student models.
\cite{park2025exploration} introduced Chemeleon, which was designed to generate chemical compositions and crystal structures from both textual descriptions and 3-dimensional structural data.
However, the vastness of the possible combinations of chemical composition makes comprehensive exploration time-consuming and computationally demanding.
Regarding explainability, an important transparency concept, \cite{xu2025technology} used \acrfull{shap} as explainable \acrshort{ai} while employing biterm topic modelling for the analysis of patents related to \acrshort{lib} research.
\acrshort{shap}, however, like other posthoc explainability methods, does not provide the level of intuitive explanations for humans like textual explanations that \acrfull{cot} reasoning from \acrshort{llm}s provide in a step-by-step fashion.



\subsection{Battery Production, Maintenance \& Sustainability}

This subsection covers findings related to the entire battery life cycle identified in Section \ref{bat_lifecycle}.
Interestingly, several deep learning methods have been explored in the diagnostics of batteries, including \acrshort{soc} estimation \citep{bian2024exploring}.
Adequate monitoring, diagnosis and maintenance ensure the prevention of battery overcharge, undue discharge, and explosion, thereby extending battery life \citep{liu2022review}.
\cite{bian2024exploring} explored \acrshort{soc} estimation for \acrshort{lib} using \acrshort{llm}s.
The task can be challenging due to harsh temperatures and dynamic operations, which are just a couple of factors out of several others that challenge reliable estimates within a module.
In their work, they proposed hard prompt generator to translate \acrshort{lib} data into instruction and answer text while a soft prompt encodes task-specific information of various \acrshort{lib}s into a set of independent vectors.
Since datasets are important for data-driven battery property predictions and estimates, some work focused on this.
\cite{huang2020database} introduced an automatically-generated materials dataset from the literature by extracting battery materials and 5 functional properties using ChemDataExtractor, which is a toolkit with \acrshort{nlp} techniques for materials science.
The extracted functional properties are capacity, conductivity, Coulombic efficiency, energy density, and voltage.
\cite{shon2023extracting} extracted the ionic conductivities of \acrshort{sse} from scientific literature to create a materials database.
Meanwhile, \cite{el2021can} noted that the lack of standard benchmark and how electrode and cell properties are reported in the domain are some of the challenges in \acrshort{lib} research giving rise to issues of lack of reproducibility among others.

Furthermore, \cite{oprea2025unveiling} rightly identified that there are  environmental concerns over the production of batteries, their use and disposal while  \cite{ren2025automated} addressed the disassembly of retired electric vehicle battery packs in their work by proposing disassembly-oriented knowledge graph.
Deep learning methods face challenges such as poor generalization and robustness \citep{bian2024exploring}.
This is complicated at harsh sub-zero ambient temperatures when \acrshort{soc} drops drastically with fluctuations due to chemical reactions in the \acrshort{lib}s.
Besides, to use language modeling for \acrshort{soc} estimation, there is the need for models to learn measurements in numerical format, though this is a typical challenge \citep{jiang2025applications}.

\section{Discussion}
\label{discussion}

The application of \acrshort{nlp} in the battery domain is growing.
However, challenges exist, as already highlighted.
For example, from Table \ref{sota_results}, we can observe that \acrshort{nlp} coverage across the battery life cycle stages is uneven.
Below, we further highlight some of the key challenges and subsequently discuss details of our proposed conceptual \acrshort{tlp} framework that may be capable of addressing some of the challenges.

\subsection{Challenges}

Although our work utilized 3 reputable databases, the number of final relevant papers is relatively limited.
In the future, increasing the number of databases may be helpful, though this is not guaranteed, since papers overlap across databases.
Meanwhile, from the review, missing in Table \ref{sota_results} is agentic \acrshort{ai} for any of the tasks, which may be because agentic \acrshort{ai} that is based on \acrshort{llm}s, is relatively recent.
This is clearly a limitation in existing attempts, which our framework addresses.
Also missing is any task related to the new EU \acrshort{dbp} or \acrfull{dpp}, which may also be because it is a relatively new proposal.
Among some of the challenges in battery research are the lack of standard battery benchmark data, proprietary data, vocabulary standards, and interoperability of tools \citep{clark2022toward}.
Different authors used different self-created datasets.
As a result, the \acrshort{sota} comparison made by \cite{choi2024accelerating} in their work may be considered unfair because their results were compared with other works which evaluated different datasets.  
Also, it would have been helpful if many of the authors gave unique names to their datasets for easy identification.
Many of these shortcomings have negatively affected transparency in the field.


\subsection{\acrshort{tlp} framework for \acrshort{dbp} and battery predictions}

\acrshort{ai} transparency provides an avenue for responsible human oversight.
The new EU \acrshort{dbp}, which is an electronic record that may be stored in the cloud and is part of the \acrshort{dpp}, provides the required transparency along the entire battery life cycle and promotes a circular economy \citep{popowicz2025digital,rizos2024implementing}.
The passport contains battery-specific texts, numeric values and dates.  
The initiative is similar to the concepts of data card, data statement, or model card in \acrshort{nlp} \citep{pushkarna2022data,bender2018data,mitchell2019model}.
Data cards or statements provide summaries of \acrshort{ml} data in a structured way with explanations of the processes and rationale behind the data.
The \acrshort{dbp} also aligns with the \acrfull{fair} guiding principles for scientific data management, which formulates a guideline for those who want to enhance the reusability of their data, and provides machines with the ability to automatically find and use such data \citep{wilkinson2016fair}.
It does not include personal data of the stakeholders, who include end-users, manufacturers, recyclers, and regulatory bodies, among others  \citep{berger2022digital}.

Just as it is beneficial for \acrshort{llm}s to have detailed prompts or instructions for improved performance \citep{10.1007/978-3-031-70442-0_5}, it is beneficial to provide as much detail as necessary in the digital passport for the benefit of \acrshort{ai} systems, besides the stakeholders.
Consequently, \acrshort{dbp} information can be transformed for \acrshort{tlp} or \acrshort{bms} modeling and predictions since a \acrshort{bms} is responsible for monitoring and managing a battery pack for safe and optimal performance \citep{shen2019review}.
The \acrshort{dbp} will provide information on material origin, composition, chemical substances, carbon footprint, capacity, hazardous substances, \acrfull{soh}, battery status (e.g. reused), number of charging and discharging cycles, performance, recycling and disposal aspects, among others.
\acrshort{bms}s typically provide information on the \acrshort{soh} and the expected battery 
lifetime \citep{berger2022digital}.

\begin{figure}[h!]
\centering
\includegraphics[width=0.95\textwidth]{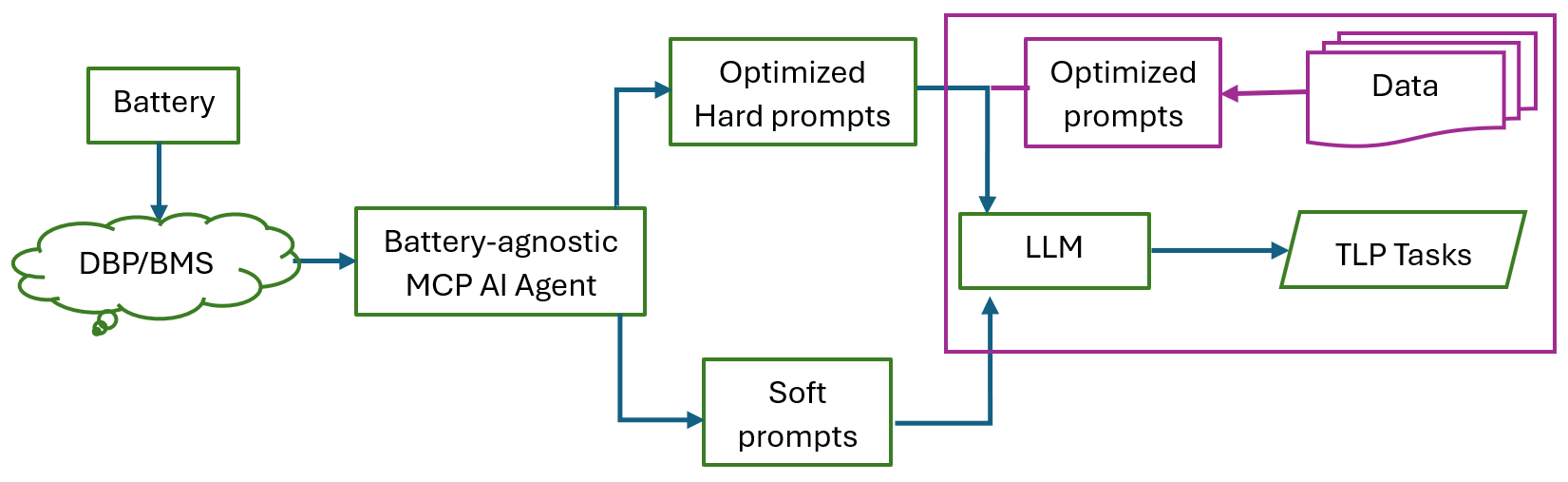}
\caption{Our conceptual \acrshort{tlp} framework for \acrshort{dbp} and battery-related predictions, based on hard-soft prompts \citep{bian2024exploring,shen2019review,NEURIPS2023_a0054803}.}
\label{tlp_framework}
\end{figure}

We propose a simple conceptual \acrshort{tlp} framework for battery predictions, including various tasks identified in Table \ref{sota_results}.
Figure \ref{tlp_framework} depicts the framework.
The information provided by the \acrshort{dbp} or \acrshort{bms} will be useful within the framework.
It involves a battery-agnostic \acrfull{mcp} \acrshort{ai} agent that can connect to external tools that have information of the \acrshort{dbp} or \acrshort{bms} and provides soft prompts (continuous feature vectors learned by prompt-tuning), which will be combined with optimized hard prompts (plain text inputs enhanced with gradient-based optimization) \citep{NEURIPS2023_a0054803}.
The combination will then be supplied as input to a capable multimodal \acrshort{llm} for relevant \acrshort{tlp} task predictions, e.g. \acrshort{soc} estimation or \acrshort{ir}.
In addition, the framework has a component capable of being a standalone part (indicated in the purple box of Figure \ref{tlp_framework}), which can perform tasks related to traditional datasets.
The strategy of combining both hard and soft prompts (or prompt parameters) can be beneficial by providing complementary perspectives as detailed and precise input to the \acrshort{llm}. 
The benefit of using \acrshort{llm}s for the tasks (instead of traditional methods) is that they have emergent properties and the capability to perform coding and other \acrshort{ml} tasks that can support predictions.
One disadvantage with dense \acrshort{llm}s is their size but using a \acrfull{moe} can be useful for mitigating this because not all the parameters need to be activated during inference.
The framework is novel because, in addition to some of the earlier points, the combination of its parts and application to \acrshort{dbp}s are new.


\section{Conclusion}
\label{conclusion}

The application of \acrshort{nlp} along the entire battery life cycle is growing, with increasing research, reports, logs and other types of documents. 
This work presented a systematic survey in the field, describing some traditional \acrshort{nlp} tasks and emerging ones, as they relate to the battery domain.
We showed, with the battery life cycle, that \acrshort{nlp} can play important parts at different stages, from materials sourcing to recycling and repurposing.
Our findings, within its limitations and possible omission of relevant studies, revealed that challenges still exist, especially with standardizing the benchmarks, creating the right evaluation framework or metric for materials discovery, and employing even more modern \acrshort{nlp} methods, such as agentic \acrshort{ai}.
The EU \acrshort{dbp} is another innovation not addressed in any of the papers surveyed.
Hence, we introduced a novel conceptual \acrshort{tlp} framework for battery predictions to address some of the challenges, though the framework will need validation in future work.
With the planned \acrshort{dbp}, the conceptual \acrshort{tlp} framework will have even more data for driving potentially better predictions.
The inclusion of multimodal \acrshort{ai} (featuring text, image, and sensor data) may also provide improved performance along the battery value chain.

\ack{This work is partly supported by the Wallenberg AI, Autonomous Systems and Software Program (WASP).
ChatGPT-5 was used only to generate the images in Figure \ref{batt_prod} for illustrative purposes.
This study is registered with a protocol on open science framework (OSF) at https://osf.io/rgcuq/overview.
As a limitation, the report did not include conflict resolution or inter-rater agreement, however, different aspects of the papers reviewed were synthesized into the paper by the authors.
The consequence of these limitations is the likelihood of a reduced possibility of total reproducibility of this work.
However, we included what is required to reproduce this work by adding artifacts for verification.
The data from this work is available at https://github.com/LTU-Machine-Learning/battery\_survey.
Furthermore, we tested Google Scholar by using the following search terms and Boolean operator OR to compare with our original search result: “text mining” OR “information extraction” OR “big language models” OR “knowledge extraction” AND “battery”.
Based on ranking selection, 8 out of 9 (about 90\%) of the results on the first page were the exact same papers in our original results.
This reduced to about 60\% over the first 5 pages, as we noticed paper relevance and paper fit reduced, as expected, since relevance to any search reduces as one goes down the ranking.}

\section*{Conflict of Interest}
All authors declare no conflicts of interest.


\appendix

\section{Search links}
\label{appsec1}

Google scholar\textsuperscript{2}: \\
https://scholar.google.com/scholar?q=text+AND+\%22language+processing\%22+AND+battery \&hl=en\&as\_sdt=0\%2C5\&as\_ylo=2017\&as\_yhi=2025

\acrshort{ieeex}: 

https://ieeexplore.ieee.org/search/searchresult.jsp?action=search\&newsearch=true\&matchBoolean=true\&queryText= (\%22All\%20Metadata\%22:text)\%20AND\%20(\%22All\%20Metadata\%22:\%22language\%20processing\%22) \%20AND\%20(\%22All\%20Metadata\%22:battery)\&highlight=true\&returnFacets=ALL\&returnType=SEARCH \&matchPubs=true\&rowsPerPage=50\&pageNumber=1

Scopus: \\
https://www.scopus.com/results/results.uri?sort=plf-f\&src=s\&sid=87cba225c81851a1f3193513dccf661b\&sot =a\&sdt=a\&sl=80\&s=text+AND+\%22language+processing\%22+AND+battery+AND+PUBYEAR+\%3E+2016+AND+
PUBYEAR+\%3C+2026\&origin=savedSearchNewOnly\&txGid=f0e09e9bcf4ff3781509c6543c2c2898\&sessionSearchId=
87cba225c81851a1f3193513dccf661b\&limit=10

\section{List of non-relevant papers}

Below, we identify the non-relevant papers in the search results by listing DatabasePaperNumber (and the focus of the paper).

\textbf{Google scholar} G14 (about electrical equipment malfunction not battery), G16 (about consumer sentiments), G20 (neurocognitive assessment batteries - a set of clinical equipment), G21 (materials of inorganic glasses), G24 (Encryption in Mobile Ad Hoc Network), G26 (Quaternion algebra), G27 (Skateboard Monitoring Device), G29 (technology lexical database), G30 (requirement management in engineering), G32 (vehicle diagnostics using free-text customer service reports), G33 (Crowdfunding Videos), G34 (Android-Based Text Extraction), G35 (Materials Applications), G36 (in-text citation analysis), G39 (Analysis of Software Industry), G41 (neural databases), G42 (L2 Listening Proficiency), G43 (ensemble learning), G44 (Video summarization), G45 (Corporate Sustainability Reports ), G46 (systems to detect significant future business changes), G48 (Unstructured Data from Medical Reports), G50 (Detection of Cognitive Decline), G51 (\acrfull{sa} for movie reviews), G52 (Digital shop floor management), G53 (Plastic Waste Recycling), G54 (Aviation Safety Reports), G55 (ADVANCES IN Computer  Science), G58 (proceedings document), G59 (corrosion-resistant alloy), G60 (Monitoring Alzheimer’s Disease), G61 (Fault Diagnosis of Signal Equipment), G62 (Language impairment in adults), G63 (clinical neuropsychology), G64 (properties of alloys), G65 (design research), G66 (therapist facilitative interpersonal skills), G67 (Loneliness in Older Adults), G68 (exploring GPT-3), G69=G71 (mild cognitive impairment), G70=G72 (Semantic Web), G73 (fMRI Dataset), G74 (sarcasm), G76 (product reviews), G77 (Personality and Psychological Distress), G78 (general language processing), G79 (Customer Satisfaction), G80 (Cognitive Functions), G81 (Clinical Data Generation), G82 (Psycholinguistic Assessments), G83 (opportunity discovery), G84 (IoT for aquaculture), G85 (Human-Machine Interaction), G86 (Arabic Sentiment Analysis), G87 (Symptom Documentation), G88 (Software Test Case Generation), G90 (extractive text summarization), G91 (cognitive plausibility), G92 (extraction from polymer literature), G93 (morphemic boundaries), G94 (Arabic \acrshort{nlp}), G95 (Arabic Text Categorization), G96 (personnel selection), G97 (Generative \acrshort{ai}), G98 (organisational culture), G99 (Energy Districts), G101 (Complex Engineered Systems), G102 (Legal Informatics), G103 (Customers' Sentiment Analysis), G104 (Measurement Extraction), G105 (IoT Based Voice Assistant), G106 (deep learning for \acrshort{nlp}), G107 (Geolocation Context), G108 (Eye movements), G109 (Grid Monitoring), G110 (neuroimaging), G111 (psychological constructs), G112 (Examination System), G113 (Depression Disorder), G114 (Sentiment Classification), G115 (RAMS Information for Metro Vehicles), G116 (Neurodevelopmental Disorders:), G117 (Readers with Autism), G118 (Aspect Sentiment Analysis).\\

\textbf{\acrshort{ieeex}} I2 (Supply Chain Information), I3 (Short Text Clustering), I4 (general natural language processing), I5 ( Aphasia Speech), I7 (Computer Crimes Safety of IOT Networks), I8 (Software Energy Consumption), I9 (edge devices), I10 (Sentiment Polarity Computation), I11 (Comparative Sentiment Analysis), I12 (News Articles Summarization), I13 (Correcting Typing Errors), I14 (sentiments around aspects), I15 (Spinal Cord Stimulator), I16 (Unveiling Comment Insights), I17 (geotemporal visualization of Twitter analysis), I20 (Cross-Lingual Topic Model ), I21 ( opinion mining), I22 (Opinion Mining via Intrinsic and Extrinsic Domain Relevance), I23 (Aspect-Based Opinion Mining), I24 (Hospital Assisting Multitasking System), I25 (Agriculture Monitoring), I28 (fuzzy domain sentiment ontology), I29 (Mobile Cloud Support ), I30 (general Generative AI), I31 (Speech Signal Quality Assessment ), I32=I33=134=135=136=137=138 (Practical Guide to Machine Learning, NLP, and Generative AI), 
\\

\textbf{Scopus} S4 (Arabic text summarization ), S6 (medical text classification), S8 (general 	
priors-augmented retrieval 
priors-augmented retrieval 
Industrial applications of \acrshort{llm}s), S9 (digital matching methods), S10 ( properties of crystalline materials), S11 (stress detection), S14 (hydrogen supply), S15 (perovskite solar devices), S16 (dementia), S17 (ophthalmic diseases), S20 (metal-organic frameworks), S21 (spin-orbit torque materials), S22 (TAP rule security detection), S23 (sentence representation of GitHub), S24 (technologies in aviation), S25 (electrocatalytic hydrogen evolution), S26 (speech-in-noise hearing tests), S27 (cognitive impairment), S28 (clinical review generation), S30 (patients with stroke), S31 (schizophrenia, bipolar disorder, and depression), S32 (patients with schizophrenia), S33 (EEG data for differential diagnosis), S34 (mechanical parts), S35 (Alzheimer’s disease), S36 (imidazolium-based ionic liquids), S37 (collaborative robots), S38 (drug-target binding affinity), S39 (Williams syndrome), S40 (thermal CO2 hydrogenation), S41 (Vocal Course ), S42 (subsidy policy on new energy vehicles ), S44 (immunogenicity), S45 (dyslexia intervention), S47 (Aircraft EWIS safety risk), S48 (waste management policy), S50 (computational literature review), S51 (Cost Outcome Pathway), S52 (media discourse), S54 (molecular representation), S56 (open-source Agile practices), S57 (Fiber-Reinforced Composites ), S58 (Photovoltaic forecasting), S59 (polymer processing database), S60 (metal-organic frameworks), S61 (electronic health records), S62 (Malware detection), S63 (Mongolian Emotional Speech Synthesis ), S64 (hydropower projects ), S65 (human–robot collaboration), S66 (multidimensional information extraction),  S68 (electricity load forecasting), S69 (priors-augmented retrieval), S70 (atypical power system failures), S71 (hyperbolic soliton families), S72 (electroretinogram signal generation), S74 (category frameworks), S75 (Process Systems Engineering), S76 (Neural tracking of continuous speech), S77 (Emotion Vocabulary Recognition), S83 (tropical flood susceptibility), S84 (Emotional Speech Synthesis),  S85 (fuzzy technology trajectories), S86 (AI-autonomous robotics), S87 (bigrams and promising patents), S89 ( IT Professional Skills), S93 (Power Sample Feature Migration ), S94 (Electric Vehicle Charging), S95 (disentanglement, selection, and reaggregation), S96 (fake review detection), S98 (competence-based HRM), S99 (Task Crowdsourcing), S101 (speech in noise), S102 (consumers’ personalized preferences), S103 (Transcranial Direct Current Stimulation), S104 (Hard-of-Hearing Children), S105 (supercapacitor researches), S106 (Management of Skin Conditions), S107 (Approaches for 3D-Printing), S108 (blood pressure estimation), S109 (Digital Health Interventions ), S110 (Tools for Scientific Review Writing),

\section{Quality of papers (based on publication venues)}
\label{qual_appendix}

\begin{table}[h]
\footnotesize
\caption{Quality of publications for Google Scholar based on venue and sorted in descending order of SCImago Journal Rank (SJR) (\footnotesize Not all venues have SJR \& since it covers only Scholar, it is a weak proxy for risk of bias.)\label{quality_papers}}
\begin{tabular}{p{0.08\linewidth} p{0.51\linewidth} p{0.06\linewidth}}
\toprule
 \textbf{Paper}   & \textbf{Venue} & \textbf{SJR} \\
\midrule
 G2$_r$ & Joule & 12.922 \\
 G22 & Advanced Functional Materials & 5.022\\
 G23 & Nature Communications & 4.904 \\
 G7 & Green Energy and Intelligent Transportation & 3.375 \\

 G4, G9 & Communication Materials   & 2.136 \\
 G56 & Energy & 2.06 \\
 G5 & Chemical Science & 2.011 \\
 G40$_r$ & Expert Systems with Applications & 1.939 \\
 G8 & Scientific Data  & 1.907 \\
 G31 & Scientific Data & 1.901 \\
 G11 & Cell Reports Physical Science &  1.784\\
 G75 & Machine Learning and Knowledge Extraction & 1.548 \\
 G57 & Machine Learning with Applications & 1.225 \\
 G13 & Scientific Reports &  0.893 \\

 G10 & Electrochemistry Communications  & 0.766 \\
 G49 & International journal of energy research & 0.755 \\

 G1 & Journal of Chemical Information and Modeling & 1.43 \\
 G18 & IETE Technical Review  & 0.585 \\
 G3$_p$ &  Computational Materials &  \\
 G6, G12 & arXiv &  \\
 G15$_r$ & Technologies  &  \\
 G17 & Apollo  &  \\
 G19 & iTech SECOM & \\
 G25 & SAE Mobulus & \\
 G28 & IEEE Xplore & \\
 G37$_r$ & Asia Information Retrieval Societies Conference & \\
 G38 & ACS Applied Materials \& Interfaces & \\
 G47$_r$ & International Council on Systems Engineering & \\
 G89 & Empirical Methods in Natural Language Processing &  \\
 G100 & ACS Sustainable Chemistry \& Engineering & \\
\bottomrule
\end{tabular}
\end{table}

\bibliographystyle{wb_stat}
\bibliography{WileySTAT}

\end{document}